\documentclass[journal]{IEEEtran}

\ifCLASSINFOpdf
\else
   \usepackage[dvips]{graphicx}
\fi
\usepackage{url}

\usepackage{graphicx}
\usepackage{hyperref}
\usepackage{amsmath}
\usepackage{amssymb}
\usepackage{mathtools}
\usepackage{adjustbox}
\usepackage{lipsum}
\usepackage{algorithm}
\usepackage{graphicx}
\usepackage{textcomp}
\usepackage{mathtools}
\usepackage{adjustbox}
\usepackage{lipsum}
\DeclareMathOperator*{\argmin}{argmin} 
\usepackage{algpseudocode}
\usepackage{expl3}
\usepackage{float}

\begin{document}

\title{Estimation of a Causal Directed Acyclic Graph Process using Non-Gaussianity}

\author{Aref Einizade and Sepideh Hajipour Sardouie
\thanks{Aref Einizade, and Sepideh Hajipour Sardouie (Corresponding author, email: hajipour@sharif.edu) are with the Department of Electrical Engineering, Sharif University of Technology, Tehran, Iran.}}

\markboth{Journal of \LaTeX\ Class Files}
{Shell \MakeLowercase{\textit{et al.}}: Bare Demo of IEEEtran.cls for IEEE Journals}
\maketitle

\begin{abstract}

Numerous approaches have been proposed to discover causal dependencies in machine learning and data mining; among them, the state-of-the-art VAR-LiNGAM (short for Vector Auto-Regressive Linear Non-Gaussian Acyclic Model) is a desirable approach to reveal \textit{both} the instantaneous and time-lagged relationships. However, all the obtained VAR matrices need to be analyzed to infer the final causal graph, leading to a rise in the number of parameters. To address this issue, we propose the CGP-LiNGAM (short for Causal Graph Process-LiNGAM), which has significantly fewer model parameters and deals with only \textit{one} causal graph for interpreting the causal relations by exploiting Graph Signal Processing (GSP).  
\end{abstract}

\begin{IEEEkeywords}
Causal Discovery, Graph Signal Processing (GSP), Causal Graph Process (CGP), Linear Non-Gaussian Acyclic Model (LiNGAM), Directed Acyclic Graph (DAG).
\end{IEEEkeywords}

\IEEEpeerreviewmaketitle

\section{Introduction}

Causal Discovery (CD) \cite{pearl2009causality,peters2017elements} of time-series data, also known as Dynamic CD (DCD), has gained significant attention due to its capability to reveal causal relationships between observational data. A wide variety of DCD methods have been proposed based on different points of view. For a comprehensive survey, please refer to \cite{assaad2022survey} and the references therein. Two popular and well-known types of causal dependency order are instantaneous and time-lagged ones which are mostly addressed by Structural Equation Model (SEM) \cite{peters2017elements}, and Vector Auto-Regressive (VAR) ones \cite{peters2017elements}. Among the SEM-based approaches, the popular Linear Non-Gaussian Acyclic Model (LiNGAM) addresses the identifiability issue by assuming the exogenous disturbances are non-Gaussian \cite{shimizu2006linear}. Besides, the state-of-the-art VAR-LiNGAM method \cite{hyvarinen2010estimation} has been proposed to reveal both the instantaneous and time-lagged dependencies. As well as desirable advantages of the VAR and VAR-LiNGAM, having a high number of free model parameters and the need for interpreting all VAR matrix coefficients to infer the final causal graph are still major drawbacks \cite{mei2016signal}.

On the other hand, in most recording systems of time-series, the interacting sensors, such as brain regions \cite{ji2021survey}, geographic temperature locations \cite{mei2016signal}, etc., can be considered the nodes of a meaningful underlying (and possibly directed) causal graph. For example, the learned graphs from brain signals can reveal the directed interactions between latent brain sources \cite{hyvarinen2010estimation}. These graphs are rarely known in real machine learning applications and need to be carefully learned/estimated from observational data \cite{assaad2022survey,ji2021survey}. Building upon this causal graph, the different recorded measurements on different nodes (in \textit{one} time index) can be interpreted as a graph signal, which provides utilizing Graph Signal Processing (GSP) tools \cite{ortega2018graph,dong2020graph,mei2016signal,ortega2022introduction} in different areas such as image \cite{liu2018graph,cheung2018graph,yan2020depth} and video \cite{mondal2021moving,giraldo2021emerging} processing.

Recently, a GSP-based method named Causal Graph Process (CGP) \cite{mei2016signal} has been proposed to address the mentioned drawbacks of the VAR by modeling the VAR coefficient matrices with graph polynomial filters; however, this model can not model the instantaneous dependencies. To address the mentioned issues, in the proposed (graph) shift invariant \cite{mei2016signal} CGP-LiNGAM analysis, the discovery of causal relationships relies on only \textit{one} and \textit{shared} underlying Directed Acyclic Graph (DAG) \cite{mei2016signal} to reveal both the instantaneous and time-lagged dependencies, unlike almost all of the other VAR-based approaches, such as VAR-LiNGAM \cite{hyvarinen2010estimation} and DYNOTEARS \cite{zheng2018dags,zheng2020learning,pamfil2020dynotears}, which need analyzing all the obtained VAR matrix coefficients to infer the underlying causal graph. The main contributions of the present paper are summarized as:

\begin{itemize}
    \item The proposed CGP-LiNGAM reveal both the instantaneous and time-lagged dependencies, and, is related to only \textit{one} and \textit{shared} underlying Directed Acyclic Graph (DAG) \cite{mei2016signal} (significantly fewer parameters, based on the analysis presented in Section (\ref{free_param})) unlike the state-of-the-art VAR-LiNGAM \cite{hyvarinen2010estimation} and DYNOTEARS \cite{pamfil2020dynotears}.
    
    \item We present closed-form iteration-based solutions in Section \ref{Sec3}, which facilitate applicability and reproducibility.
    
    \item Non-Gaussianity of the exogenous disturbances guarantees the identifiability of the proposed model \cite{hyvarinen2010estimation}.
    
    \item Based on the experimental analysis in Sections \ref{Resuts}-A, \ref{Resuts}-B and \ref{Resuts}-C, the CGP-LiNGAM performs more accurately and robust in causal graph recovery and prediction tasks compared to the VAR-LiNGAM \cite{hyvarinen2010estimation}, and DYNOTEARS \cite{pamfil2020dynotears}.
    
    \item Based on the additional presented analysis in Section \ref{Resuts}-D, the CGP-LiNGAM method is also robust in the case of not known CGP (VAR) order \cite{ljung1998system}. 
\end{itemize}

The notations \((.)^T\), \(\otimes\), \((.)^{\dagger}\), \( \| . \| _p\), and \(\hat{x}\) stand for transpose operator, Kronecker product, Moore-Penrose pseudo-inverse of a matrix, the \(p\)-norm of a vector, and the estimation of the true entity \(x\), respectively. 

\section{Related Work and Background}
\label{Sec2}
The observation matrix \(\textbf{X}=(\textbf{x}(0),\textbf{x}(1),...,\textbf{x}(K-1))\in\mathbb{R}^{N\times K}\) is provided to us by recording \(K\)-length signals on \(N\) sensors/nodes. In the following, the related work/background is briefly reviewed as follows:

\textbf{LiNGAM.} In an (instantaneous) SEM (\ref{SEM}), assuming the exogenous disturbance \(\textbf{e}\) is non-Gaussian, the underlying DAG \(\textbf{A}\) can be uniquely recovered by the LiNGAM analysis \cite{shimizu2006linear}:

\begin{equation}
\label{SEM}
    \textbf{x}(k)=\textbf{A}\textbf{x}(k)+\textbf{e}(k)
\end{equation}

\textbf{VAR.} The VAR modeling of time-series \(\textbf{X}\) is defined as:

\begin{equation}
\label{VAR}
    \textbf{x}(k) = \sum_{i=1}^{M}{\textbf{A}^{(i)}\textbf{x}(k-i)}+\textbf{e}(k)
\end{equation}

\noindent where \(M\) denotes the VAR-order, and \(\{\textbf{A}^{(i)}\}_{i=1}^{M}\) are VAR coefficient matrices \cite{hyvarinen2010estimation}. From the Granger \cite{granger1969investigating} analysis point of view, to recover the underlying causal graph \(\textbf{A}'\) modeling the causal relationships, one considers \(\textbf{A}'_{ij}=0\), if \(\textbf{A}_{ij}^{(k)}=0\) for all \(k\) \cite{bolstad2011causal,mei2016signal}, implying the conditional independence between the time-series at each node described by a Markov Random Field (MRF) with adjacency structure \(\textbf{A}'\) \cite{mei2016signal}.

\textbf{VAR-LiNGAM.} In this method, both the instantaneous and time-lagged dependencies are modeled using the combination of (\ref{SEM}) and (\ref{VAR}) as \cite{hyvarinen2010estimation}:

\begin{equation}
\label{VAR_LiNGAM}
    \textbf{x}(k) = \sum_{i=0}^{M}{\textbf{A}^{(i)}\textbf{x}(k-i)}+\textbf{e}(k)
\end{equation}

In this method, \(\textbf{A}^{(0)}\) is considered a DAG, and the exogenous disturbance \(\textbf{e}\) is assumed to be non-Gaussian to guarantee the identifiablity \cite{hyvarinen2010estimation}. Then, similar to the VAR model (\ref{VAR}), the sparsity pattern of all \(\{\textbf{A}^{(i)}\}_{i=0}^{M}\) must be analysed to recover the true underlying causal graph \(\textbf{A}'\) \cite{hyvarinen2010estimation}.

\textbf{GSP background}. A (possibly directed) graph \(\mathcal{G}=(\textbf{V},\textbf{A})\) is characterized by its vertex set \(\textbf{V}=\{\mathop{v}_1,...,\mathop{v}_N\}\) and its adjacency matrix \(\textbf{A}\in\mathbb{R}^{N\times N}\). If \(\textbf{A}_{ij}\ne{0}\), the \(i\)th node is a parent of the \(j\)th node, and the \(j\)th node is a descendant of the \(i\)th node \cite{peters2017elements}. We call a node \textit{pure parent}, if it is not the descendant of any other nodes. A graph signal \(\textbf{x}=(x_1,...,x_N)^T\in\mathbb{R}^{N\times1}\) is a mapping \(\textbf{x}:\textbf{V}\rightarrow \mathbb{R}\), assigning the \(i\)th vertex the value of \(x_i\) \cite{ortega2018graph}. A \(L\)-order graph polynomial filter is also defined as \(c(\textbf{A})=c_0\textbf{I}+c_1\textbf{A}+...+c_L\textbf{A}^L\) \cite{mei2016signal}, where \(\textbf{I}\) denotes the identity matrix of size \(N\).

\textbf{Causal Graph Process (CGP).} An \(M\)-order CGP model has the following form \cite{mei2016signal}:

\begin{equation}
\label{CausalGraphDepp}
    \textbf{x}(k) = \sum_{i=1}^{M}{P_i(\textbf{A},\textbf{c})\textbf{x}(k-i)} + \textbf{e}(k)
\end{equation}

\noindent where \(P_i(\textbf{A},\textbf{c})=\sum_{j=0}^{i}{c_{ij}\textbf{A}^j}\) and \(\textbf{c}\) collects the scalar polynomial coefficients as (\(1\le i \le M\) and \(0\le j \le i\)):

\begin{equation}
    \textbf{c}=(c_{10}, c_{11},...,c_{ij},...,c_{MM})^T
\end{equation}

\section{The Proposed CGP-LiNGAM Approach}
\label{Sec3}

In our CGP-LiNGAM framework, the instantaneous (LiNGAM) and time-lagged (CGP) parts share \textit{one} underlying DAG \(\textbf{A}\). To recover both these causal dependencies, our (graph) shift invariant \cite{marques2017stationary} proposed framework is shown as:

\begin{equation}
\label{CGPLiNGAM1}
        \textbf{x}(k) =\textbf{A}\textbf{x}(k)+ \sum_{i=1}^{M}{\overbrace{P_i(\textbf{A},\textbf{c})}^{\textbf{R}_i}\textbf{x}(k-i)}+\textbf{e}(k)
\end{equation}

The proposed model (\ref{CGPLiNGAM1}) can be rewritten as:

\begin{equation}
\label{CGPLiNGAM2}
    \textbf{x}(k) = \sum_{i=1}^{M}{\tilde{\textbf{R}}_{i}\textbf{x}(k-i)} + \tilde{\textbf{e}}(k)
\end{equation}

\noindent where the following theorem helps to recover \(\tilde{\textbf{R}}_{i}\) in (\ref{CGPLiNGAM2}), and

\begin{equation}
\begin{split}
    & \tilde{\textbf{R}}_{i} = (\textbf{I}-\textbf{A})^{-1}\textbf{R}_{i};\:\: \tilde{\textbf{e}}(k) = (\textbf{I}-\textbf{A})^{-1}\textbf{e}(k)
\end{split}
\end{equation}


\textbf{Theorem 1:} \textit{In the model (\ref{CGPLiNGAM2}), \(\tilde{\textbf{R}}_{i}\) and \(\tilde{\textbf{R}}_{j}\) commute, i.e., \(\tilde{\textbf{R}}_{i}\tilde{\textbf{R}}_{j}=\tilde{\textbf{R}}_{j}\tilde{\textbf{R}}_{i}\), for \(i,j=1,2,...,N\) under the conditions:}

\noindent\textit{A1) \textbf{A} is a DAG.}

\noindent\textit{A2) \(\forall i:\: |\lambda_i(\textbf{A})|<1\), where \(\lambda_i(\textbf{A})\) is the \(i\)th eigenvalue of \(\textbf{A}\).}

\textit{Proof:} If \textbf{A} is a DAG, \(\textbf{I}-\textbf{A}\) is invertible \cite{shimizu2006linear,hyvarinen2010estimation}, and if \(\forall i:\: |\lambda_i(\textbf{A})|<1\), the the infinite sum of the geometric series \(\sum_{r=0}^{\infty}{\textbf{A}^{r}}\) is summable, and equal to \((\textbf{I}-\textbf{A})^{-1}\) as \cite{szidarovszky2002introduction}:

\begin{equation}
    (\textbf{I}-\textbf{A})^{-1} = \sum_{r=0}^{\infty}{\textbf{A}^{r}}
\end{equation}

Then, the VAR coefficient matrices \(\{\tilde{\textbf{R}}_{i}\}_{i=1}^M\) in (\ref{CGPLiNGAM2}) commute because they can be as (infinite) graph polynomial filters:

\begin{equation}
\begin{split}
&\tilde{\textbf{R}}_{i}= (\textbf{I}-\textbf{A})^{-1}\textbf{R}_{i}= \left[\sum_{r=0}^{\infty}{\textbf{A}^{r}}\right]\left[\sum_{j=0}^{i}{c_{ij}\textbf{A}^j}\right]\\
&=\sum_{r=0}^{\infty}{\sum_{j=0}^{i}{c_{ij}\textbf{A}^{r+j}}}.\:\:\:\:\:\:\:\:\:\:\:\:\:\:\: \blacksquare
\end{split} 
\label{graphFilter}
\end{equation}

In the following, the proposed three-step approach for recovering \(\{\tilde{\textbf{R}}_i\}_{i=1}^M\), \(\textbf{A}\), \(\{\textbf{R}_i\}_{i=1}^M\), and \(\textbf{c}\) is presented.

\subsection{Recovering \(\{\tilde{\textbf{R}}_i\}_{i=1}^M\)}

Based on Theorem 1, the commutativity terms are included in the following multi-convex optimization \cite{mei2016signal} to obtain the (possibly sparse) graph polynomial filters \(\left\{\tilde{\textbf{R}}_{i}\right\}_{i=1}^M\) as:

\begin{equation}
\label{SolveR1}
\begin{split}
    & \hat{\tilde{\textbf{R}}}_i=\argmin_{\tilde{\textbf{R}}_{i}}{\frac{1}{2}\sum_{k=M}^{K-1}{\left\|\textbf{x}(k)-\sum_{i=1}^{M}{\tilde{\textbf{R}}_i\textbf{x}(k-i)}\right\|_2^2}+\lambda_1\|vec(\tilde{\textbf{R}}_1)\|_1}\\    
    & \text{Subject to:\:\:\:} \tilde{\textbf{R}}_i\tilde{\textbf{R}}_j=\tilde{\textbf{R}}_j\tilde{\textbf{R}}_i;\:\:\:i,j=1,...,N
\end{split}
\end{equation}

An alternating optimization to (\ref{SolveR1}) is expressed as:

\begin{equation}
\label{SolveR2}
\begin{split}
&\hat{\tilde{\textbf{R}}}_i=\argmin_{\tilde{\textbf{R}}_i}{\frac{1}{2}\sum_{k=M}^{K-1}{\left\|\textbf{x}(k)-\sum_{i=1}^{M}{\tilde{\textbf{R}}_i\textbf{x}(k-i)}\right\|_2^2}}\\
& +\lambda_1\|vec(\tilde{\textbf{R}}_1)\|_1+\lambda_3\sum_{i\ne{j} }{\|\tilde{\textbf{R}}_i\tilde{\textbf{R}}_j-\tilde{\textbf{R}}_j\tilde{\textbf{R}}_i\|_F^2}   
\end{split}
\end{equation}

\noindent where, with the definitions \(\textbf{X}_m = (\textbf{x}(m),\:\textbf{x}(m+1),\:...,\:\textbf{x}(m+K-M-1))\) and \(\tilde{\textbf{r}}_i=vec(\tilde{\textbf{R}}_i)\), (\ref{SolveR2}) can be rewritten as (with the details \cite{petersen2008matrix} in the Appendix section):



\begin{equation}
\label{Solve_ri}
\begin{split}
    \hat{\tilde{\textbf{r}}}_{i} = \argmin_{\tilde{\textbf{r}}_{i}}{\|\boldsymbol{\Psi}_i\tilde{\textbf{r}}_i-\tilde{\textbf{y}}_i\|_2^2+\lambda_1\|\tilde{\textbf{r}}_1\|_1}
\end{split}
\end{equation}

\noindent where \(\textbf{0}\) is a all-zero array with appropriate size, and


\begin{equation}
\begin{split}
& \boldsymbol{\Psi}_{i} =\\
& \left(\frac{\sqrt{2}}{2}\textbf{B}_i^T, \sqrt{\lambda_3}\boldsymbol{\Phi}_1^T, \hdots, \sqrt{\lambda_3}\boldsymbol{\Phi}_{i-1}^T, \sqrt{\lambda_3}\boldsymbol{\Phi}_{i+1}^T, \hdots, \sqrt{\lambda_3}\boldsymbol{\Phi}_M^T\right)^T\\
\end{split}
\end{equation}


\begin{equation}
\begin{split}
& \tilde{\textbf{y}}_{i} =\left(\frac{\sqrt{2}}{2}\textbf{y}_i^T, \textbf{0}^T, \hdots, \textbf{0}^T\right)^T\\
\end{split}
\end{equation}


\begin{equation}
\begin{split}
& \textbf{B}_i = \textbf{X}^T_{M-i}\otimes \textbf{I}, \:\:\:\boldsymbol{\Phi}_i=\textbf{R}^T_i\otimes \textbf{I} - \textbf{I}^T \otimes \textbf{R}_i,\\
& \textbf{y}_i = vec(\textbf{X}_M)-vec\left(\sum_{j=1\ne{i}}^{M}{\tilde{\textbf{R}}_j\textbf{X}_{M-j}}\right) 
\end{split}
\end{equation}

Afterwards, the \textit{unique} (identifiable) \cite{mei2016signal} closed-form solutions to (\ref{SolveR2}) are expressed as:

\begin{equation}
\label{sol_ri}
    \hat{\tilde{\textbf{r}}}_{i\ne{1}}=\boldsymbol{\Psi}_i^{\dagger}\tilde{\textbf{y}}_i,\:\:\:i=2,...,M
\end{equation}

\noindent and \(\hat{\tilde{\textbf{r}}}_1\) can be obtained using standard \(\mathop{l}\)-1 minimization approaches in (\ref{Solve_ri}), such as LASSO \cite{tibshirani1996regression}. 

\subsection{Recovering \(\textbf{A}\)}

By recovering \(\{\tilde{\textbf{R}}_i\}_{i=1}^M\), the residual \(\tilde{\textbf{e}}\) is also obtained from (\ref{CGPLiNGAM2}), and the relation \(\tilde{\textbf{e}}(k) = (\textbf{I}-\textbf{A})^{-1}\textbf{e}(k)\) can be rewritten:

\begin{equation}
\label{LiNGAM_eq}
  \tilde{\textbf{e}}(k) = \textbf{A}\tilde{\textbf{e}}(k) + \textbf{e}(k)  
\end{equation}

\noindent which is an SEM, and, due to the non-Gaussianity of the exogenous disturbance \({\textbf{e}}\), the underlying DAG \(\textbf{A}\) can be \textit{uniquely} (identifiable) inferred using the LiNGAM analysis \cite{shimizu2006linear,hyvarinen2010estimation} on \(\tilde{\textbf{e}}\) exploiting FastICA \cite{hyvarinen1999fast} approach. 

\subsection{Recovering \(\{\textbf{R}_i\}_{i=1}^M\) and \(\textbf{c}\)}

By recovering \(\{\tilde{\textbf{R}}_i\}_{i=1}^M\) and \(\textbf{A}\), the graph polynomial filters \(\{\textbf{R}_i=(\textbf{I}-\textbf{A})\tilde{\textbf{R}}_{i}\}_{i=1}^M\) are obtained, and, inspiring from the CGP model \cite{mei2016signal} (i.e., \(\textbf{R}_i=P_i(\textbf{A},\textbf{c})=\sum_{j=0}^{i}{c_{ij}\textbf{A}^j}\)), the polynomial coefficients \(\textbf{c}\) are estimated by minimizing the following convex optimization using standard \(\mathop{l}_1\)-regularized least squares methods, such as LASSO \cite{tibshirani1996regression}:

\begin{equation}
\label{solve_c}
    \hat{\textbf{c}}_i = \argmin_{\textbf{c}_i}{\frac{1}{2}\left\|vec(\hat{{\textbf{R}}}_i)-\textbf{Q}_i\textbf{c}_i\right\|_2^2+\lambda_2\|\textbf{c}_i\|_1}
\end{equation}

\noindent where \(\textbf{Q}_i=\left(vec(\textbf{I}),\:vec(\hat{\textbf{A}}),...,\:vec(\hat{\textbf{A}}^i)\right)\), and \(\textbf{c}_i=(c_{i0},c_{i1},...,c_{ii})^T\) \cite{mei2016signal} for \(i=1,...,M\). Concisely, our proposed CGP-LinGAM algorithm is summarized in Algorithm \ref{alg:cap}. 

\algnewcommand\INPUT{\item[\textbf{Input:}]}%
\algnewcommand\OUTPUT{\item[\textbf{Output:}]}%

\begin{algorithm}[!t]
\caption{: CGP-LiNGAM}\label{alg:cap}
\begin{algorithmic}[1]
\INPUT $\textbf{X}\in\mathbb{R}^{N\times T},\: M,\: \{\lambda_i\}_{i=1}^3$
\OUTPUT Causal DAG $\textbf{A}\in\mathbb{R}^{N\times N}$, Polynomial coefficients $\textbf{c}$
\State Initialize: \(t=1\), \(\hat{\textbf{R}}^{(0)}=\textbf{0}\) 
\While{Convergence}
\For{$i=1:M$}
\State Estimate \(\tilde{\textbf{R}}^{(t)}_i\) by fixing \(\{\tilde{\textbf{R}}^{(t-1)}_j\}_{j=1\ne{i}}^{M}\), and using

\(\:\:\:\:\) CVX \cite{grant2008cvx} in (\ref{SolveR1}) or (\ref{SolveR2}), or using LASSO\cite{tibshirani1996regression}/ 

\(\:\:\:\:\:\:\)closed-form solutions in (\ref{sol_ri})
\EndFor
\State $t\leftarrow t+1$
\EndWhile{: return \(\{\tilde{\textbf{R}}_{i}\}_{i=1}^M\) and \(\tilde{\textbf{e}}\) in (\ref{CGPLiNGAM2}})
\State  Estimate \(\textbf{R}_0=\textbf{A}\) using LiNGAM analysis \cite{shimizu2006linear} on \(\tilde{\textbf{e}}\)
\State Obtain the causal effect matrices \(\{\textbf{R}_i=(\textbf{I}-\textbf{A})\tilde{\textbf{R}}_{i}\}_{i=1}^M\)
\State Solve for \(\textbf{c}\) in (\ref{solve_c}) by \(\mathop{l}\)-1 minimization, e.g., LASSO \cite{tibshirani1996regression}
\end{algorithmic}
\end{algorithm}

\subsection{Number of learnable model parameters}
\label{free_param}
In the CGP-LiNGAM (\ref{CGPLiNGAM1}), the underlying DAG \(\textbf{A}\) has \(\frac{N(N-1)}{2}\) free parameters (because of the technically being lower triangular), and \(\textbf{c}\in\mathbb{R}^{\frac{M(M+3)}{2}}\), compared to the VAR-LiNGAM \cite{hyvarinen2010estimation} (\ref{VAR_LiNGAM}), which needs \(\frac{N(N-1)}{2} + MN^2\) free parameters to describe the model. On the other hand, in real application, usually \(N\gg M\) \cite{hyvarinen2010estimation,mei2016signal}, so the CGP-LiNGAM (with combined \(\frac{N(N-1)}{2} + \frac{M(M+3)}{2}\) free parameters) is considerably parsimonious compared to the VAR-LiNGAM. 

\subsection{Complexity Analysis of CGP-LiNGAM (Algorithm \ref{alg:cap})}
For a specific \(i\) and an convergence iteration in Line 4, optimization (\ref{SolveR2}) is naively dominated by the matrix-matrix product of \(\{\tilde{\textbf{R}}_i \tilde{\textbf{R}}_j\}_{j=1\ne{i}}^{M}\) with \(\mathcal{O}((M-1)N^3)\), and matrix-vector product of \(\{\tilde{\textbf{R}}_j\textbf{x}(k-j)\}_{k=M,j=1}^{K-1,M}\) with \(\mathcal{O}((K-M)N^2)\) complexity \cite{mei2016signal}. As a result, the overall complexity of one convergence iteration is  \(\mathcal{O}(M^2N^3+KMN^2)\). In Line 8, the computational complexity of the LiNGAM analysis typically \(\mathcal{O}(KN^3+N^4)\) \cite{shimizu2011directlingam}. The matrix-matrix products of \(\{(\textbf{I}-\textbf{A})\tilde{\textbf{R}}_i\}_{i=1}^M\) in Line 9 take total complexity of \(\mathcal{O}(MN^3)\). In Line 10, the minimization (\ref{solve_c}) for each \(i\) is dominated by multiplications \(\textbf{Q}_i^T\textbf{Q}_i\in\mathbb{R}^{(i+1)\times(i+1)}\) and \(\textbf{Q}_i^Tvec(\tilde{\textbf{R}}_i)\in\mathbb{R}^{(i+1)\times1}\), which take \(\mathcal{O}((i+1)^2)\) operations \cite{mei2016signal}, and, combined complexity of \(\mathcal{O}(M(\sum_{i=1}^{M}{(i+1)^2}))\approx\mathcal{O}(M^4)\). All in all, assuming \(M\ll K\), the total (worst-case and naive) computational complexity of Algorithm \ref{alg:cap} is approximately \(\mathcal{O}(KMN^2+KN^3+N^4+M^4)\). Note that the (possible) sparsity of the underlying DAG can severely reduce the complexity \cite{mei2016signal}. 

\begin{figure*}[!t]
    \centering
    \centerline{\includegraphics[width=20cm, trim={1cm 0.3cm 2cm 1cm}, clip=true]{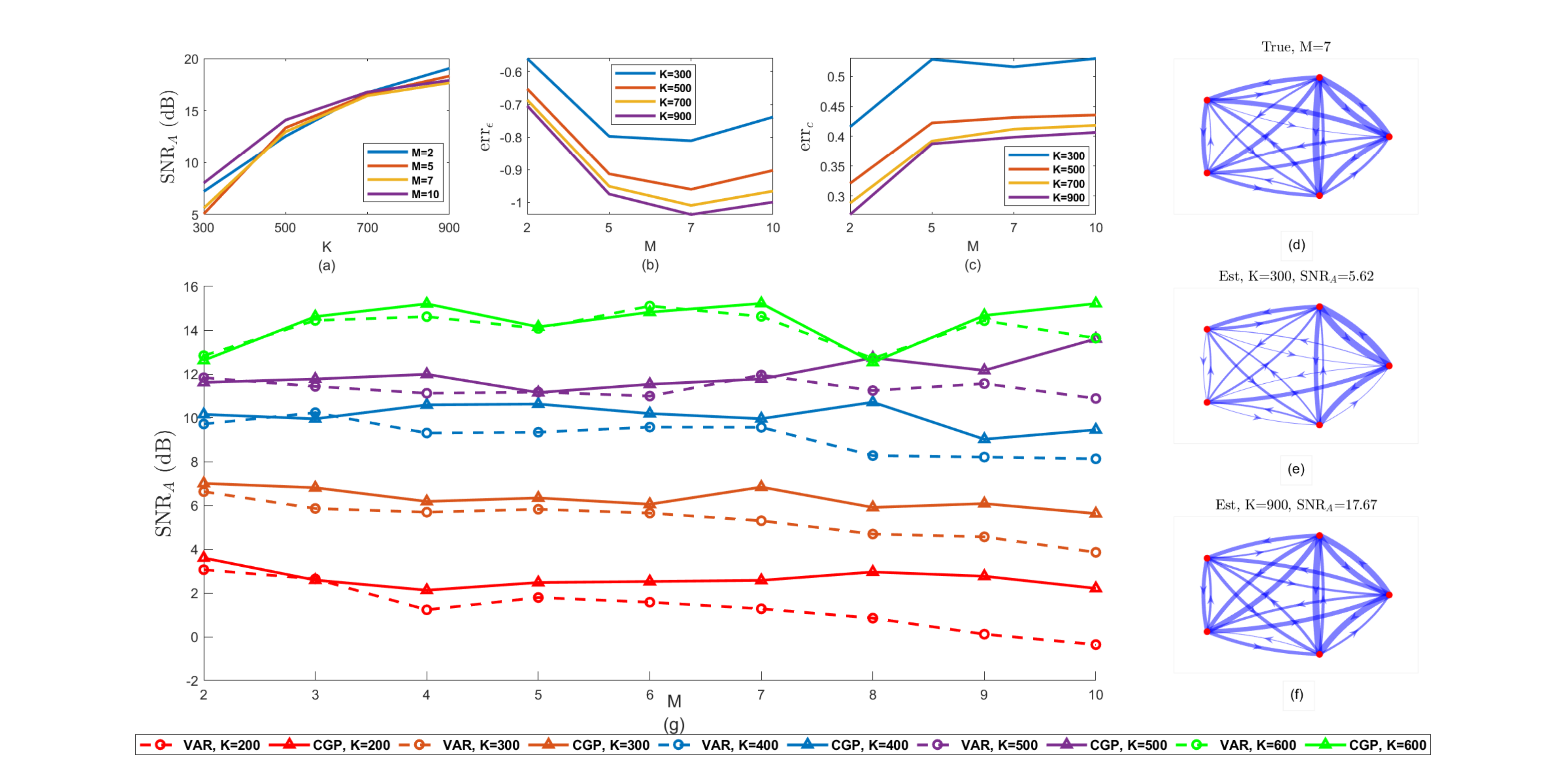}}
    \caption{(a-c) The average of the performance evaluation metrics over thirty Monte-Carlo realizations, by varying the number of samples \(K\) and the CGP-order \(M\) in the spans of \(\{300, 500, 700, 900\}\) and \(\{2, 5, 7, 10\}\), respectively. (d-f): An example average (\(M=7\)) to the graph structures (Figure \ref{Fig1}, (d:f)) over Monte-Carlo realizations. (g): Comparison of the graph recovery performance of the proposed CGP-LiNGAM method with the celebrated VAR-LiNGAM \cite{hyvarinen2010estimation}}
    \label{Fig1}
\end{figure*}


\section{Experimental Results and Discussion}
\label{Resuts}
In this section, we provide experimental analysis in two categories: 1) ablation study: analyzing the performance of the CGP-LiNGAM in different situations, 2) comparison study: comparing the performance of the CGP-LiNGAM with that of the state-of-the-art, i.e., VAR-LiNGAM \cite{hyvarinen2010estimation}. The underlying DAG \(\textbf{A}\) and also the non-Gaussian exogenous disturbance \(\textbf{e}\) are generated as described in \cite{shimizu2006linear}. Note that, to have sparse DAGs, we generate \(\textbf{A}\) (with \(N=5\)) with just \textit{one} pure parent. The generated data were divided into three parts of the train, validation (to find the optimal hyperparameters using grid search) and test (to evaluate the prediction based on the optimized model) \cite{mei2016signal}. 

\subsection{Ablation Study}

We vary the number of samples \(K\) and the CGP-order \(M\) in the spans of \(\{300, 500, 700, 900\}\) and \(\{2, 5, 7, 10\}\), respectively, to study the effect of these changes on the recovery performance. The polynomial coefficients \(\textbf{c}\) are generated as \(2^{i+j}c_{ij}\sim {0.5[\mathcal{U}(-1, -0.45)+\mathcal{U}(0.45,1)]}\) to model coefficient decay with distance to the current time sample \cite{mei2016signal}. Besides, to evaluate the quality of the recoveries, we consider the following scale-free metrics \(SNR_{\textbf{A}} = 20\log({\|\textbf{A}\|_F}/{\|\hat{\textbf{A}}-\textbf{A}\|_F})\), \(err_{\textbf{c}} = {\|\hat{\textbf{c}}-\textbf{c}\|_2}/{\|\textbf{c}\|_2}\), and




\begin{equation}
\label{err_c}
\begin{split}
    & err_{\epsilon} =  \mathop{\mathbb{E}}\left[\frac{1}{N}\|\textbf{x}(k)-f(\hat{\textbf{A}}, \hat{\textbf{c}}, \textbf{X}'_{k-1})\|_2^2\right]\\
    & -\mathop{\mathbb{E}}\left[\frac{1}{N}\|\textbf{x}(k)-f({\textbf{A}}, {\textbf{c}}, \textbf{X}'_{k-1})\|_2^2\right]
\end{split}    
\end{equation}

\noindent where \(f({\textbf{A}}, {\textbf{c}}, \textbf{X}'_{k-1})=\textbf{A}\textbf{x}(k)+\sum_{i=1}^{M}{P_i(\textbf{A},\textbf{c})\textbf{x}(k-i)}\), and \(err_{\epsilon}\) models the prediction error on the test data \cite{mei2016signal}. Also, the expectations \(\mathbb{E}\{.\}\) in (\ref{err_c}) are approximated by the sample mean over time samples. The average of the mentioned metrics and also an example average (\(M=7\)) to the graph structures over thirty Monte-Carlo realizations are illustrated in Figure \ref{Fig1} (a:c), and (d:f), respectively. Note that although the underlying true graphs are acyclic, the plotted averaged ones can have cycles. From these results, it can be admitted that, on average, the graph recovery is superior in small values of \(M\), i.e., \(M=2\), in case of having enough time samples, i.e., \(K=900\). Besides, the higher the number of the time samples \(K\), the lower the prediction error \(err_{\epsilon}\) \cite{mei2016signal}. Also, the \(err_{\textbf{c}}\) is small and robust against the changes of \(M\) if the number of time samples is not very small. i,e, \(K>300\).  
\subsection{Comparison with State-of-the-art}
\label{comparison}
In this subsection, we compare the graph recovery performance of the CGP-LiNGAM with the celebrated VAR-LiNGAM in Figure \ref{Fig1} (g) by varying \(K\) and \(M\) in the spans of \(\{200:600\}\) and \(\{2:9\}\), respectively. The polynomial coefficients \(\textbf{c}\) are generated as \(2^{i+j}c_{ij}\sim {\mathcal{N}(1, 0.01)}\). It is well illustrated in this figure that the CGP-LiNGAM is more robust and also superior compared to the VAR-LiNGAM, especially in a small number of time samples \(K\), i.e., \(K\le500\), and also in high values of causal dependency \(M\), i.e., \(M\ge5\).

\subsection{Comparison with DYNOTEARS \cite{pamfil2020dynotears}}
\begin{figure}[!t]
	\centering
	\centerline{\includegraphics[width=7.5cm, trim={0.7cm 0cm 1cm 0.5cm}, clip=true]{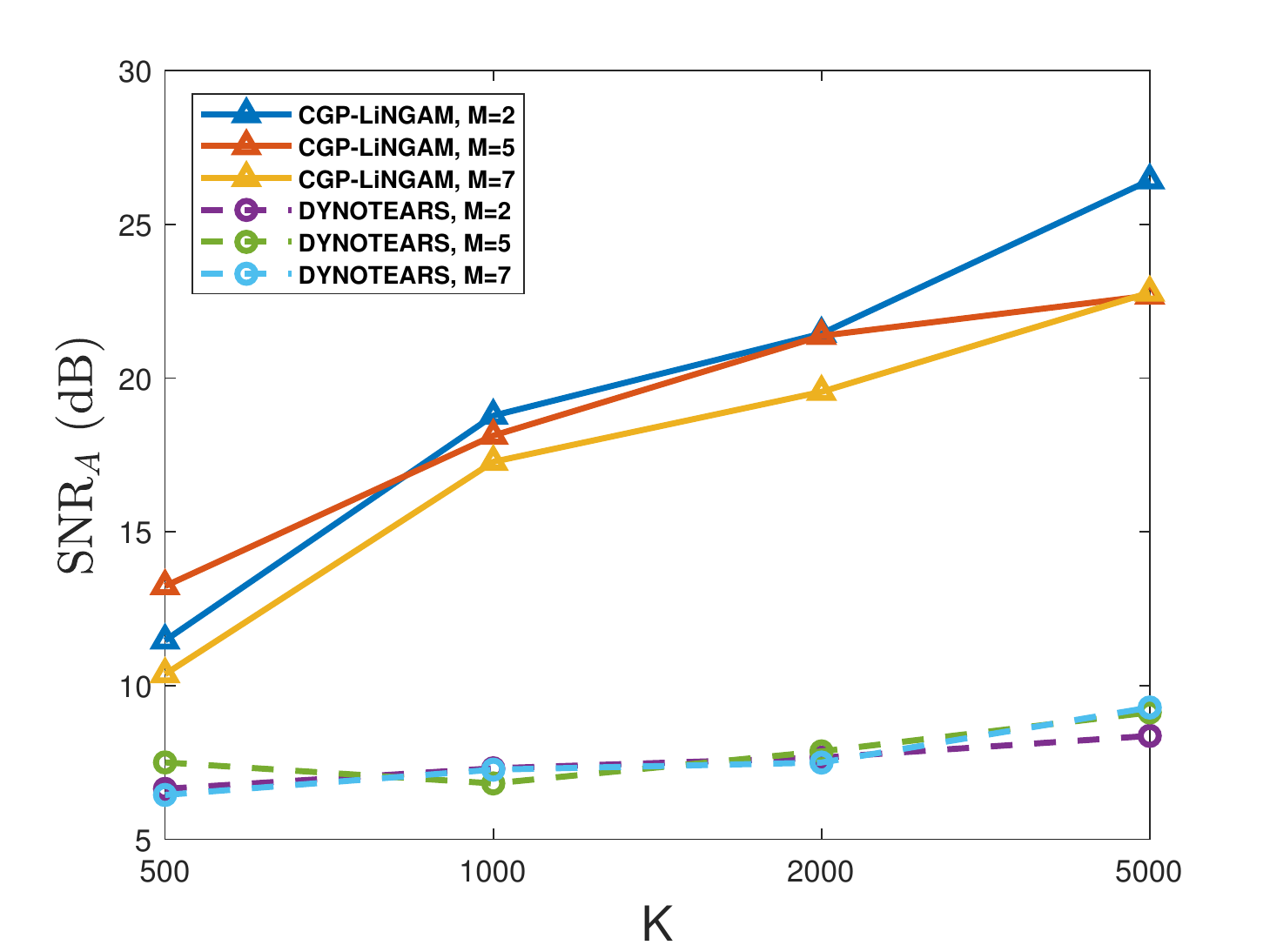}}
	\caption{Comparison of the graph recovery performance with the DYNOTEARS \cite{pamfil2020dynotears} method.}
	\label{Dynotears_fig}
\end{figure}

The graph recovery results of the CGP-LiNGAM and DYNOTEARS \cite{pamfil2020dynotears} methods in different CGP order \(M\in\{2,5,7\}\) and across time samples \(K\in\{500,1000,2000,5000\}\) are shown in Figure \ref{Dynotears_fig}. It can be seen that a similar trend of improving the graph recovery with increasing the sample size \(K\) is observed in this figure for both methods; however, the DYNOTEARS \cite{pamfil2020dynotears} has clearly failed to successfully recover the true DAGs, while the CGP-LiNGAM has estimated them with high quality. The DYNOTEARS model can be stated as \cite{pamfil2020dynotears}:


{\footnotesize {\begin{equation}
			\textbf{x}(k) =\textbf{A}\textbf{x}(k)+ \sum_{i=1}^{M}{\textbf{W}_i\textbf{x}(k-i)}+\textbf{e}(k)
\end{equation}}}

\noindent where \textit{only} \textbf{A} has been assumed to be an acyclic graph, and \(\{\textbf{W}_i\}_{i=1}^M\) are unconstrained coefficient matrices (like VAR-LiNGAM \cite{hyvarinen2010estimation}) leading to increased number of learnable parameters to \(\frac{N(N-1)}{2}+MN^2\), in comparison to the proposed method which has significantly reduced \(\frac{N(N-1)}{2}+\frac{M(M+3)}{2}\) learnable ones, because often \(M\ll N\) \cite{hyvarinen2010estimation,mei2016signal}.

\begin{figure}[!t]
	\centering
	\centerline{\includegraphics[width=6.8cm, trim={1cm 0.7cm 1cm 0.5cm}, clip=true]{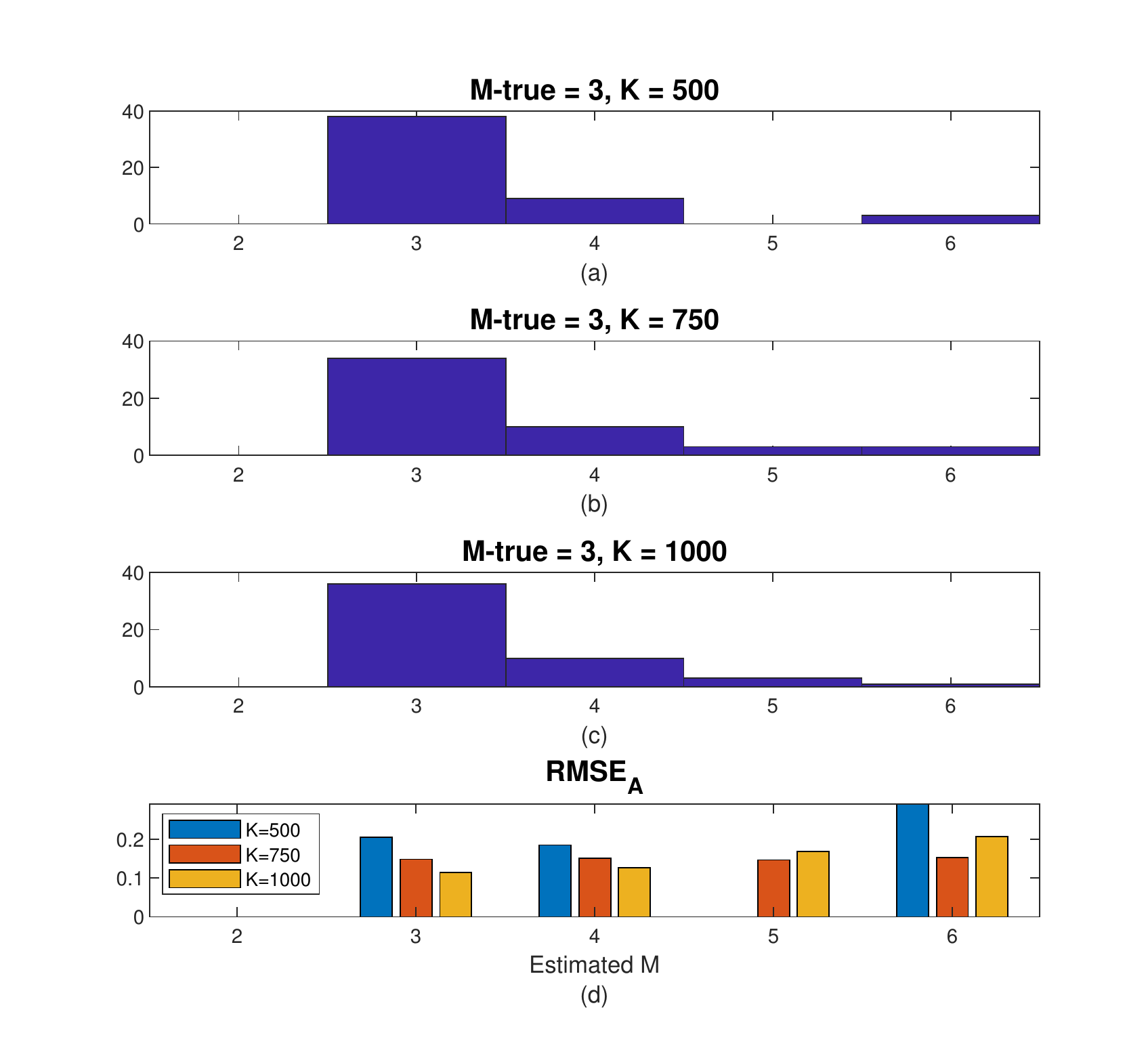}}
	\caption{(a)-(c): The histograms of the estimated \(M\), in each \(K\in\{500,750,1000\}\) based on minimum nAIC (\ref{nAIC}), (d): The averaged \(\textbf{RMSE}_\textbf{A}=\frac{\|\hat{\textbf{A}}-\textbf{A}\|_F}{\|\textbf{A}\|_F}\) over realizations of each selected \(M\).}
	\label{Fig2}
\end{figure}

\subsection{Choosing optimal CGP-order \(M\)}

In most real applications, the CGP/VAR-order \(M\) is considered a fix or a prior known \cite{mei2016signal,hyvarinen2010estimation}, however, due to some uncertainty conditions, e.g., existing high amount of noise, choosing it optimally can considerably boost the recovery/modeling performance \cite{ljung1998system}. In this way, to investigate the possibility of correct selection of \(M\), we generate fifty Monte-Carlo realizations similar to Section \ref{comparison} with CGP-order \(M_{true}=3\) and number of time samples \(K\in\{500,750,1000\}\). Then, in each \(K\), the normalized Akaike's Information Criterion (nAIC) (\ref{nAIC}) \cite{ljung1998system} is calculated in the span of \(M\in\{2:6\}\) and the CGP-order \(M\) leading to minimum nAIC is considered as the estimation of \(M_{true}\).

{\footnotesize \begin{equation}
		\label{nAIC}
		nAIC_{\hat{M}}=\log\left(\det\left(\frac{1}{K}\sum_{k=1}^{K}{\hat{\textbf{e}}(k)\hat{\textbf{e}}(k)^T}\right)\right) + \frac{2n_p}{K}
\end{equation}}

\noindent where \(det(.)\) denotes the determinant operation, \(\hat{\textbf{e}}\) is the estimated disturbances corresponding to \(\hat{M}\) and the number of the estimated parameters \(n_p=\frac{N(N-1)}{2}+\frac{\hat{M}(\hat{M}+3)}{2}\). Figure \ref{Fig2} (a)-(c) show the histograms of the estimated \(M\), in each \(K\). These results show that, in almost every \(K\), \(M_{true}\) is successfully recovered. Moreover, in Figure \ref{Fig2} (d), the averaged \(\textbf{RMSE}_\textbf{A}=\frac{\|\hat{\textbf{A}}-\textbf{A}\|_F}{\|\textbf{A}\|_F}\) over realizations of each selected \(M\) is plotted, which shows that, even in selected wrong \(M\) cases, i.e., \(M\ne M_{opt}\), the graph recovery remains fairly robust. 

\section{Conclusion}
\label{Sec5}

In this paper, we proposed the CGP-LiNGAM method to reveal both the instantaneous and time-lagged causal relationships in time-series by considering the node-specific time samples as graph signals on underlying causal DAGs. Compared to the state-of-the-art VAR-LiNGAM, our method has significantly fewer parameters, deals with only \textit{one} underlying causal graph, and performs more robust against a low number of time samples and a high degree of causal dependency. Extending the CGP-LiNGAM to the time-varying graphs \cite{kalofolias2017learning,giraldo2022reconstruction} is our future research direction.

\section{Appendix (Simplifications of eq. (\ref{SolveR2})}
\label{Appendix}

Due to the convexity of (\ref{SolveR2}) w.r.t \(\tilde{\textbf{R}}_i\) (by fixing \(\{\tilde{\textbf{R}}_{j\ne i}\}_{j=1}^{M}\)) \cite{mei2016signal}, and with the definitions \(\textbf{X}_m = (\textbf{x}(m),\:\textbf{x}(m+1),\:...,\:\textbf{x}(m+K-M-1))\) and \(\tilde{\textbf{r}}_i=vec(\tilde{\textbf{R}}_i)\), the closed-form solutions can be obtained as:

\begin{equation}
\begin{split}
\hat{\tilde{\textbf{R}}}_{i} & =\argmin_{\tilde{\textbf{R}}_{i}}{\frac{1}{2}\left\|\textbf{X}_M-\sum_{j=1}^{M}{\tilde{\textbf{R}}_j\textbf{X}_{M-j}}\right\|_F^2+\lambda_1\|vec(\tilde{\textbf{R}}_1)\|_1}\\
& +\lambda_3\sum_{j\ne{i} }{\left\|[\tilde{\textbf{R}}_i, \tilde{\textbf{R}}_j]\right\|_F^2}  \\
& = \argmin_{\tilde{\textbf{R}}_{i}}{\frac{1}{2}\left\|vec\left(\tilde{\textbf{R}}_i\textbf{X}_{M-i}\right)-\overbrace{vec\left(\textbf{X}_M-\sum_{j=1\ne{i}}^{M}{\tilde{\textbf{R}}_j\textbf{X}_{M-j}}\right)}^{\textbf{y}_i}\right\|_2^2}\\
& +\lambda_1\|vec(\tilde{\textbf{R}}_1)\|_1 +\lambda_3\sum_{j\ne{i} }{\left\|vec\left(\tilde{\textbf{R}}_i \tilde{\textbf{R}}_j\right)-vec\left(\tilde{\textbf{R}}_j \tilde{\textbf{R}}_i\right)\right\|_2^2}\\
\end{split}
\end{equation}


\noindent 
Then, using the relation \(vec(\mathcal{A}\mathcal{B}\mathcal{C})=(\mathcal{C}^T\otimes\mathcal{A})vec(\mathcal{B})\) \cite{petersen2008matrix}:

\begin{equation}
\begin{split}
\hat{\tilde{\textbf{r}}}_{i} & =\argmin_{\tilde{\textbf{r}}_{i}}{\frac{1}{2}\left\|\overbrace{(\textbf{X}_{M-i}^T\otimes \textbf{I})}^{\textbf{B}_i}\tilde{\textbf{r}}_i-\textbf{y}_i\right\|_2^2+\lambda_1\|\tilde{\textbf{r}}_1\|_1}\\
& +\lambda_3\sum_{j\ne{i} }{\left\|\overbrace{(\tilde{\textbf{R}}_j^T\otimes \textbf{I}-\textbf{I}^T\otimes\tilde{\textbf{R}}_j )}^{\boldsymbol{\Phi}_j}\tilde{\textbf{r}}_i\right\|_2^2}  \\
& =\argmin_{\tilde{\textbf{r}}_{i}}{\|\boldsymbol{\Psi}_i\tilde{\textbf{r}}_i-\tilde{\textbf{y}}_i\|_2^2+\lambda_1\|\tilde{\textbf{r}}_1\|_1}
\end{split}
\end{equation}

\bibliographystyle{unsrt}
\bibliography{References}

\end{document}